\documentclass{article}

\usepackage{PRIMEarxiv}
\usepackage{multirow}
\usepackage[utf8]{inputenc} 
\usepackage[T1]{fontenc}    
\usepackage{hyperref}       
\usepackage{url}            
\usepackage{booktabs}       
\usepackage{amsfonts}       
\usepackage{nicefrac}       
\usepackage{microtype}      
\usepackage{lipsum}
\usepackage{fancyhdr}       
\usepackage{graphicx}       
\graphicspath{{media/}}
\usepackage{amsmath}
\usepackage[table,xcdraw]{xcolor}

\title{EDSNet: Efficient-DSNet for Video Summarization
}

\author{
  Ashish Prasad  \hspace{1cm} Pranav Jeevan \hspace{1cm} Amit Sethi \\
  Department of Electrical Engineering \\
  Indian Institute of Technology Bombay \\
  Mumbai, India  \\
  \texttt{\{21d180009, pjeevan, asethi\}@iitb.ac.in} \\
}

\begin{document}
\maketitle

\begin{abstract}
Current video summarization methods largely rely on transformer-based architectures, which, due to their quadratic complexity, require substantial computational resources. In this work, we address these inefficiencies by enhancing the Direct-to-Summarize Network (DSNet) with more resource-efficient token mixing mechanisms. We show that replacing traditional attention with alternatives like Fourier, Wavelet transforms, and Nyströmformer improves efficiency and performance. Furthermore, we explore various pooling strategies within the Regional Proposal Network, including ROI pooling, Fast Fourier Transform pooling, and flat pooling. Our experimental results on TVSum and SumMe datasets demonstrate that these modifications significantly reduce computational costs while maintaining competitive summarization performance. Thus, our work offers a more scalable solution for video summarization tasks.
\end{abstract}



\section{Introduction}

As of June 2022, more than 500 hours of video are uploaded to YouTube every minute, marking a 40\% increase from 2014~\cite{statista_youtube_2024}. This vast and largely unannotated video data underscores the increasing importance of video summarization. Video summarization involves extracting the most crucial information from a video. This technique has several applications, including managing information overload, content indexing, enhancing searchability \cite{christel2006evaluation}, social media monitoring and analysis \cite{rani2020social}, surveillance and security \cite{muhammad2020efficient, zhang2016context}, and personalized content recommendations.

A significant portion of research in supervised video summarization uses transformer encoder blocks~\cite{vaswani2017attention}, which struggle with the $O(n^2)$ complexity of self-attention, making it difficult to handle long sequences. While feasible for small-scale applications, this becomes impractical for the massive data volumes on social media, surveillance footage, and streaming platforms. To tackle this, we incorporate Nyströmformer~\cite{xiong2021nystromformer} and FNet blocks~\cite{lee2021fnet}, which reduce complexity, enabling more efficient handling of large-scale video data.

Current research in video summarization uses a frame-wise classification approach, labeling each frame as relevant or irrelevant. However, this does not reflect how humans process videos—we first understand the global context before focusing on specific moments. Our approach mimics this by using efficient token-mixers to grasp the overall plot, followed by a temporal region proposal network to identify key segments for summarization. This method involves binary classification for segment selection and offset refinement through regression, capturing global context with token-mixers and refining finer details with the regression block for accurate summarization.

\begin{figure}[htbp]
\centerline{\includegraphics[width=0.5\linewidth]{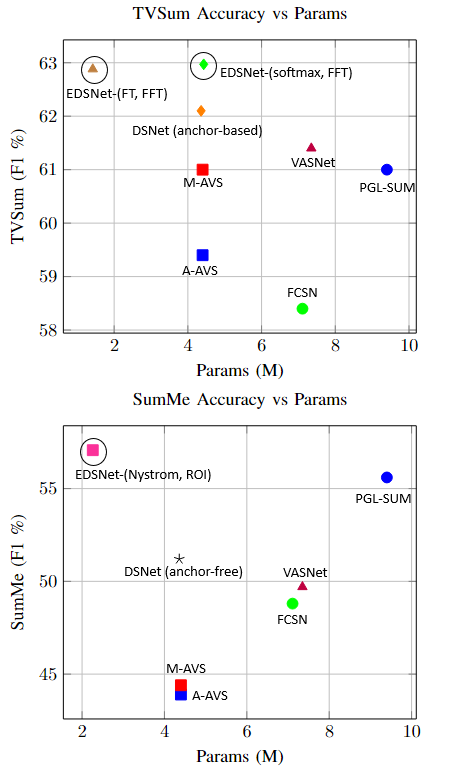}}
\caption{Plot comparing model accuracy (F1 \%) versus number of parameters for TVSum and SumMe datastes shows that EDSNet models outperform others while remaining parameter efficient. EDSNet models are circled.}
   \label{fig:model comaprision}
\end{figure}

\begin{figure*}[htbp]
\centerline{\includegraphics[width=0.9\linewidth]{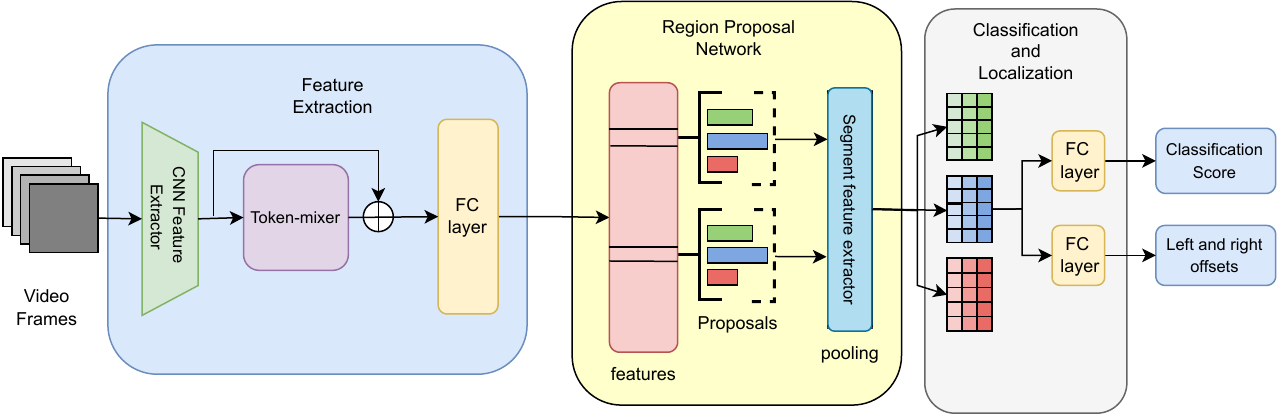}}
\caption{The model architecture of EDSNet illustrates the video summarization process, starting with a CNN feature extractor and a token-mixer for feature extraction. The outputs are refined using a fully connected layer, followed by region proposal generation and segment feature extraction. Finally, classification and localization are performed through fully connected layers to provide classification scores and segment boundary offsets, enabling accurate summarization and temporal localization of important video segments.}
   \label{fig:dsnet}
\end{figure*}

\section{Related Work}
\label{sec:related_work}

Supervised video summarization approaches focus on training models with annotated datasets to generate summaries close to human-created ones. The Fully Convolutional Sequence Network (FCSN)~\cite{rochan2018video} was an early deep learning method that used convolutional layers to encode temporal dependencies, predicting frame-level importance scores. To improve temporal modeling, the Visual-Temporal Attention-based Network (VASNet)~\cite{fajtl2019summarizing} introduced a soft attention mechanism, capturing both local and global dependencies and achieving state-of-the-art performance by effectively learning contextual frame importance. More recent approaches have incorporated advanced attention mechanisms to enhance video summarization quality. The Deep Reinforcement Learning-based Deep Summarization Network (DR-DSN)~\cite{zhou2018deep} used a reinforcement learning framework to capture long-term dependencies and contextual information. The Memory Augmented Video Summarizer (MAVS)~\cite{feng2018extractive} introduced an external memory network to store visual information from the entire video, improving the model's ability to generate comprehensive summaries.

Efficient transformers have been developed to reduce the quadratic complexity of traditional self-attention, especially for long-sequence tasks. The Nystr{\"o}mformer~\cite{xiong2021nystromformer} approximates self-attention using the Nystr{\"o}m method, enabling linear complexity for longer sequences. Linformer reduces costs through low-rank factorization~\cite{wang2020linformer}, while Performer~\cite{choromanski2020rethinking} uses kernel-based approximations for linear time complexity. The Longformer~\cite{beltagy2020longformer} combines global and local sparse attentions to handle lengthy texts efficiently. Hybrid models incorporating these efficient mechanisms with convolutional layers perform well in resource-constrained vision tasks~\cite{9706874}.

Temporal segment localization focuses on identifying the start and end times of actions in videos. Early methods, such as sliding window-based approaches~\cite{teutsch2015robust, lian2022sliding, yuan2009speeding}, used fixed-length windows to sample frames, capturing temporal dependencies but suffered from high computational costs. Recent methods leverage deep learning for more efficient localization. The convolutional-deconvolutional network~\cite{shou2017cdc} enhances boundary accuracy through temporal upsampling and spatial downsampling, while the Segment-Tube detectorr~\cite{wang2018segment} refines localization with per-frame masks. Multi-Stage CNNs~\cite{shou2016temporal} generate proposals more efficiently, and approaches like super-voxels~\cite{jain2014action} and actionness scores~\cite{yu2015fast} focus on generating action tubelets. Deep Action Proposals (DAPs)~\cite{escorcia2016daps}, utilizing LSTM networks, highlight the significance of temporal context for precise localization.

\section{Approach}

We take the Detect-to-Summarize Network (DSNet) ~\cite{zhu2020dsnet} architecture and modify the feature extraction and region proposal networks to enhance its efficiency and performance. We employed different token-mixing modules for temporal modeling and compared them on accuracy (F1 score), GPU usage, and model size. 


\subsection{Feature Extraction}

We used GoogLeNet \cite{szegedy2015going} for spatial feature extraction from video frames similar to DSNet \cite{zhu2020dsnet}. Given a video with $N$ frames, the extracted features are $v_{i}$, where $i \in \{1,2,...,N\}$. To efficiently extract temporally relevant spatial information, we replace softmax self-attention \cite{vaswani2017attention} with other token mixers.

 \textbf{Fourier transform}: The fourier transform replaces the self-attention mechanism with two 1-D Discrete Fourier Transform (DFT) along the sequence and embedding dimensions as used in FNet\cite{lee2021fnet}. The DFT decomposes sequences into their frequency components, efficiently mixing tokens without learnable parameters. The DFT operation makes the computation faster than softmax attention for longer sequences.
 

\textbf{Nystr{\"o}mformer \cite{xiong2021nystromformer}}:
Nystr{\"o}mformer approximates the standard self-attention mechanism using the Nystr{\"o}m method-based low-rank matrix approximation. By decomposing the attention matrix into smaller matrices, Nystr{\"o}mformer reduces the complexity to $O(N)$. This method preserves global context while reducing memory usage and computational overhead, making it suitable for longer sequences.

\begin{figure}[h]  
    \centering  
    \includegraphics[width=0.4\textwidth]{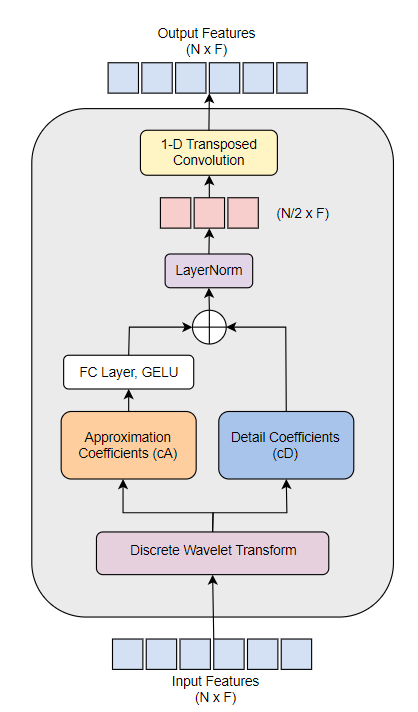}  
    \caption{DWT token-mixer module uses the 1-D DWT for token-mixing in video frame feature extraction, decomposing inputs into approximation and detail coefficients. It employs normalization, and 1D-transposed convolutions to stabilize training and refine temporal resolution. $N$ is the number of frames and $F$ is the feature dimension}
    \label{fig 4example}  
\end{figure}

\textbf{Discrete wavelet transform~\cite{jeevan2022wavemix}}: Similar to WaveMix in computer vision, we employ a 1-dimensional discrete wavelet transform (1D-DWT) for token-mixing in video summarization tasks, effectively capturing both temporal and frequency domain information. The DWT token-mixing module uses a specified wavelet (Haar) to decompose the input sequence into approximation and detail coefficients as shown in Fig.~\ref{fig 4example}. The approximation coefficients are then passed through fully connected layers with a GELU non-linearity. The output is then combined with detail coefficients components and is normalized using layer norm to stabilize training and improve convergence. A 1-D transposed convolutional layer is employed to restore sequence length after downsampling by 1-DWT, refining the temporal resolution. The DWT-based approach offers computational efficiency while capturing essential features without introducing any trainable parameters.

\subsection{Region Proposal Network}

Similar to DSNet~\cite{zhu2020dsnet}, we employ an anchor-based method for region proposals in video frames. We propose segments of lengths $l_k$ at each frame, where $k \in {1,2,…,K}$. At temporal location $t, \in {1,2,...,N}$, K interest proposals are appointed within the range 
$[t - \frac{l_{k}}{2},t + \frac{l_{k}}{2})$, where $l_k$ represents the duration of the k-th interest proposal. Thus, a total of $N \times K$ interest proposals are generated for a video sequence with $N$ frames. 

During training, we assign binary class labels (positive or negative) to interest proposals to address the class imbalance problem. Positive and negative proposals are sampled in a 1:3 ratio to alleviate the problem of class imbalance. A proposal is positive when its temporal Intersection over Union (tIoU) with any ground truth segment exceeds 0.6. A negative label is assigned if the tIoU is 0 (unimportant) or between 0 and 0.3 (incomplete). Negative samples are further divided such that unimportant and incomplete interest proposals occupy 2/3 and 1/3, respectively. We avoid assigning negative proposals with tIoU between 0.3 and 0.6, as this harms summary performance due to confusion between positive and negative proposals.

\begin{table*}[htbp]
\centering
\caption{Comparison of Different Models for Video Summarization}

\label{tab:model_comparison}
\begin{tabular}{@{} l r r rrr @{}}
\toprule
\textbf{Model} & \textbf{Params (M)} & \multicolumn{2}{c}{\textbf{Accuracy (F1 \%)}} & \multicolumn{2}{c}{\textbf{GPU Mem (MB)}} \\

& & \textbf{TVSum} & \textbf{SumMe} & \textbf{TVSum} & \textbf{SumMe}  \\ 
\midrule
A-AVS \cite{ji2019video} & 4.40 & 59.4 & 43.9 & - & -  \\
M-AVS \cite{ji2019video} & 4.40 & 61.0 & 44.4 & - & -  \\
FCSN \cite{rochan2018video} & - & 58.4 & 48.8 & - & -  \\
VASNet \cite{fajtl2019summarizing} & 7.35 & 61.4 & 49.7 & - & -  \\
DSNet (anchor-based) \cite{zhu2020dsnet} & 4.36 & 62.1 & 50.2 & 1017 & 509  \\
DSNet (anchor-free) \cite{zhu2020dsnet} & 4.36 & 61.9 & 51.2 & 1015 & 509  \\
PGL-SUM \cite{apostolidis2021combining} & 9.4 & 61.0 & 55.6 & 533 & 533  \\
MSVA \cite{ghauri2021supervised} & - & 61.5 & 53.4 & - & -  \\
MAVS\cite{feng2018extractive} & - & \textbf{67.5} & 43.1 & - & -  \\
EDSNet-(Nystrom, ROI) (SL = 12) (ours)  & \textbf{2.26} & 59.6 & \textbf{57.07} & 445 & 405  \\
EDSNet-(FT, FFT) (SL = 12,) (ours) & \textbf{1.42} & 62.88 & 48.87 & \textbf{445} & \textbf{397}  \\
EDSNet-(softmax, FFT) (SL = 4) (ours)& 4.43 & 62.97 & 49.42 & 1000 & 513  \\
\bottomrule
\end{tabular}

\end{table*}

\subsubsection{Feature Extraction for Segment Proposals} 
We replace the temporal averaged pooling layer of DSNet~\cite{zhu2020dsnet} with three different methods of pooling to extract features from segment proposals.  

\textbf{Region of interest pooling}: Region of Interest (ROI) pooling is used to manage variable-length segments by converting them into fixed-size representations suitable for fully connected layers. In our implementation, ROI pooling is applied along the temporal dimension, using average pooling for each anchor scale. However, ROI pooling's reliance on averaging can result in a loss of fine-grained details, which may not significantly impact segment classification but is crucial for accurate segment localization.

\textbf{Fast Fourier transform pooling}:
Fast Fourier transform (FFT) pooling uses FFT to retain fine-grained details that may be lost in average pooling. The Fourier transform is only applied along the temporal dimension of each segment. 

\textbf{Flat pooling}:
Flat pooling is a simpler approach where each segment is flattened directly. This method involves concatenating all segments into a single representation without any transformation.

After applying the pooling operation (except for ROI pooling), the coarse information is obtained by averaging the transformed segment along the temporal axis, while the fine-grained features are stacked together. The output dimensions change from $(N\times num\_hidden)$ to $(N \times l_k \times (num\_hidden * K))$ by flattening each segment across the temporal dimension. These features are then passed through the fully connected layer fo suitable width to change the shape to $(N \times num\_hidden)$ for further classification and regression tasks.

\begin{figure}[htbp]
\centerline{\includegraphics[width=0.5\linewidth]{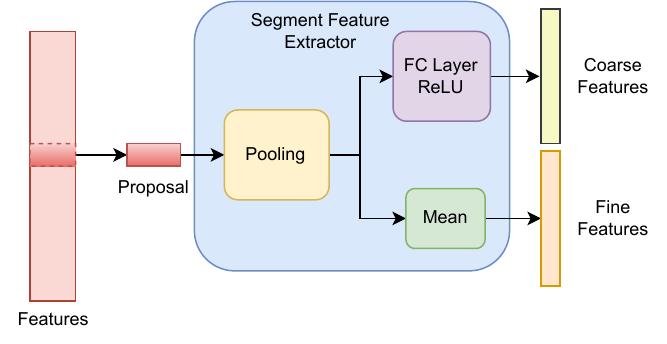}}
\caption{The segment feature extractor applies pooling operations along the temporal dimension of each segment, which is then flattened and averaged to obtain coarse features and passed through a fully connected layer with ReLU activation to extract fine-grained features.}
   \label{fig:segm}
\end{figure}

\subsection{Classification and Localization}

Similar to DSNet~\cite{zhu2020dsnet}, the pooled features are fed into the classification and regression module. The module is composed of a shared fully connected layer followed by ReLU non-linearity, layer-normalization, and two sibling output branches. The first branch outputs importance scores of proposals using coarse features (except for ROI pooling-based method), and the second branch outputs the associated center and segment length offsets using fine features (except for ROI pooling-based method). 

During testing, predicted offsets refine segment proposals, with non-maximum suppression (NMS) used to remove low-confidence and overlapping segments. To generate video summaries, we follow previous work \cite{zhou2018deep,zhang2018retrospective} where videos are first segmented into shots using Kernel Temporal Segmentation (KTS) \cite{potapov2014category}, and shot-level importance scores are calculated by averaging frame-level scores. To ensure fair comparison, shot selection is constrained to 15\% of the video length, solved as a 0/1 knapsack problem via dynamic programming to maximize the summary's importance.



\section{Experiments}
\label{sec:experiments}
\subsection{Datasets}

The datasets used in our experiments are TVSum \cite{song2015tvsum} and SumMe \cite{gygli2014creating}, two well-established benchmarks for video summarization evaluation. TVSum includes 50 videos across genres like news, how-to, documentaries, vlogs, and egocentric content, with 1,000 shot-level importance scores crowd sourced (20 per video). SumMe consists of 25 videos, each with at least 15 human-generated summaries, totaling 390 annotated summaries.

As in previous studies, we downsampled the videos to 2 frames per second (fps). Downsampling reduces computational complexity and speeds up processing while retaining sufficient visual information for effective summarization.
We employed 5-fold cross-validation with an 8:2 ratio for training and testing. The F1 score was used as the evaluation metric due to its balance between precision and recall.

\subsection{Implementation Details}
From the down-sampled Video frames, 1024-dimensional spatial image features (feature dimension size) are extracted using GoogLeNet \cite{szegedy2015going} pre-trained on ImageNet \cite{deng2009imagenet}. We use the attention mechanism to extract global attention features, which are then compressed to a 128-dimensional ($num\_hidden$) vector using a fully connected layer and ReLU Activation. A dropout of 0.5 is used. We use the same multi-task loss used by \cite{zhu2020dsnet} with the same settings of hyperparameters, and the non-maximum suppression threshold was set to 0.5. Our anchor-based model was trained for 300 epochs using the Adam optimizer, with an initial learning rate of $5 \times10^{-5}$ and a weight decay of $10^{-5}$. The experiments were conducted on the Nvidia P100 GPU available on Kaggle. GPU memory consumption is reported for a batch size of 1. 

To compare the performance, the fully connected (FC) depth was set to 1, and the F1 score was compared for various token mixers with different pooling operations and segment lengths (SL) of 4, 8, and 12.

The nomenclature for our EDSNet models is \textit{EDSNet (token-mixer, pooling)} with the name of the token-mixer in the feature extractor and pooling method used in the segmentation feature extractor.

\section{Results and Discussions}

Table~\ref{tab:model_comparison} presents a comprehensive comparison of various state-of-the-art (SOTA) models for video summarization, focusing on parameters, performance metrics, and GPU memory usage. The proposed models, EDSNet-(Nystr{\"o}m, ROI) and EDSNet-(FT, FFT), outperform several state-of-the-art (SOTA) models in terms of efficiency and accuracy. EDSNet-(Nystrom, ROI) achieves the highest accuracy on SumMe (57.07\%), surpassing PGL-SUM \cite{apostolidis2021combining} and DSNet \cite{zhu2020dsnet}. Similarly, EDSNet-(FT, FFT) and EDSNet-(softmax, FFT) deliver competitive results on TVSum (62.88\% and 62.97\%, respectively). Notably, EDSNet-(FT, FFT) has the lowest parameter count (1.42M) compared to all models. Furthermore, EDSNet-(Nystrom, ROI) demonstrates the most efficient GPU memory consumption on SumMe (405 MB), outperforming DSNet and PGL-SUM, which consume over 500 MB. Overall, our models maintain high accuracy while offering substantial improvements in resource efficiency, making them suitable for memory-constrained environments. The results of the comparison of EDSNet with different token-mixers, pooling mechanisms, and segment lengths are shown in Table \ref{tab:feature_extractor}.

\begin{table*}[h!]
\centering
\caption{Comparison of performance of EDSNet with different token-mixers, pooling types, and segment lengths on SumMe and TVSum Datasets. Green shows best and red shows worst result.}
\vspace{1mm}
\label{tab:feature_extractor}
\begin{tabular}{@{} c l r r r r c c c c @{}}
\toprule
\textbf{Segment Lengths} & \textbf{Pooling Method} & \multicolumn{4}{c}{\textbf{SumMe (F1 \%)}} & \multicolumn{4}{c}{\textbf{TVSum (F1 \%)}} \\

 & & \textbf{Nystr{\"o}m} & \textbf{Softmax} & \textbf{Fourier} & \textbf{DWT} & \textbf{Nystr{\"o}m} & \textbf{Softmax} & \textbf{Fourier} & \textbf{DWT} \\
 
\midrule


4 & FFT & \cellcolor{red!10}49.51  & \cellcolor{red!10}49.42  & \cellcolor{green!10}48.38  & \cellcolor{red!10}49.06  & \cellcolor{green!10}61.15  & \cellcolor{green!10}62.97  & 61.42  & \cellcolor{green!10}62.37  \\

 & ROI & \cellcolor{green!10}52.42  & 49.53  & 48.03  & \cellcolor{green!10}52.5  & \cellcolor{red!10}57.09  & 61.85  & \cellcolor{red!10}58.59  & \cellcolor{red!10}61.02  \\
 & Flat & 50.00  & \cellcolor{green!10}50.05 & \cellcolor{red!10}47.71 & 49.18  & 60.13 & \cellcolor{red!10}61.1  & \cellcolor{green!10}61.45 & 62.22  \\
\midrule
8 & FFT & 51.18 & 50.2  & 48.79 & \cellcolor{red!10}49.2  & \cellcolor{green!10}61.96  & \cellcolor{green!10}62.65  & \cellcolor{green!10}62.43 & \cellcolor{green!10}62.72  \\
 & ROI & \cellcolor{green!10}54.32  & \cellcolor{green!10}51.37  & \cellcolor{green!10}49.16 & \cellcolor{green!10}50.58  & \cellcolor{red!10}58.73 & 62.12  & \cellcolor{red!10}58.22 & \cellcolor{red!10}59.18  \\
 & Flat & \cellcolor{red!10}48.42  & \cellcolor{red!10}48.02  & \cellcolor{red!10}45.38 & 48.07  & 60.22 & \cellcolor{red!10}61.52  & 60.64 & 62.37  \\
\midrule
12 & FFT & 49.17 & \cellcolor{green!10}49.23  & 48.87 & 48.43 & \cellcolor{green!10}62.07 & \cellcolor{green!10}62.40  & \cellcolor{green!10}62.88 & 62.17  \\
 & ROI & \cellcolor{green!10}57.07 & 48.77  & \cellcolor{red!10}46.41 & \cellcolor{green!10}50.22  & 59.6 & \cellcolor{red!10}61.68 & \cellcolor{red!10}57.64 & \cellcolor{red!10}60.67 \\
 & Flat & \cellcolor{red!10}47.81  & \cellcolor{red!10}48.31  & \cellcolor{green!10}48.64 & \cellcolor{red!10}46.72  & \cellcolor{red!10}60.7  & 62.17  & 61.38 & \cellcolor{green!10}62.28  \\
\hline
\end{tabular}
\end{table*}

For SumMe, FFT pooling shows stable performance across different token-mixers, with Nyströmformer achieving the highest F1 scores, peaking at 51.18\% for a segment length of 8, suggesting FFT pooling effectively captures temporal features for this model. In contrast, Fourier token-mixing struggles, with a best score of 48.87\% at a segment length of 12. For TVSum, FFT pooling performs well, with softmax attention and Fourier token-mixing achieving competitive scores of 62.4\% and 62.88\%, respectively, indicating its effectiveness in handling temporal variations. ROI pooling generally boosts performance, particularly for Nyströmformer, which reaches 57.07\% and 59.6\% for SumMe and TVSum at segment length 12. Softmax attention also benefits from ROI pooling but to a lesser extent, showing it is compatible with models like Nyströmformer that rely on capturing fine-grained features. Flat pooling performs inconsistently, often yielding lower results compared to FFT and ROI, as it fails to adequately capture temporal dependencies.

\section{Ablation Studies}

\begin{figure}
    \centering
    \includegraphics[width=0.4\linewidth]{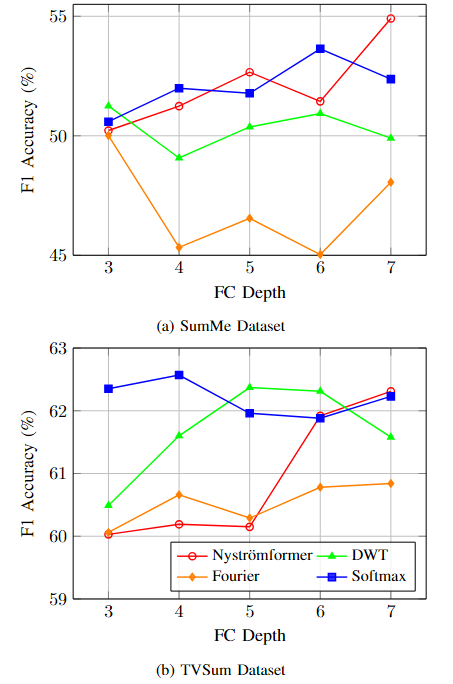}
    \caption{Comparison of Accuracy for Different token-mixing Methods at Varying FC Depths for SumMe and TVSum Datasets.}
    \label{fig:accuracy_results}
    \vspace{-4mm}
\end{figure}

\subsection{Segment Length}

At a segment length of 4, the DWT model performs well, particularly with FFT pooling, indicating that shorter segments favor models that capture both temporal and frequency domain information effectively. Nyströmformer also performs reasonably well with ROI pooling, benefiting from fine-grained temporal dependencies, while Fourier token-mixing underperforms across most settings. At a segment length of 8, Nyströmformer and Softmax attention improve, especially with ROI and FFT pooling, reaching 54.32\% and 51.37\% accuracy on SumMe, suggesting that this intermediate length balances temporal dynamics and contextual information. DWT’s advantage diminishes at this stage. At a segment length of 12, Nyströmformer excels, particularly with ROI pooling, benefiting from longer segments, while Fourier token-mixing and Softmax attention with flat pooling continue to show lower performance, indicating they struggle with longer sequences.

\subsection{Fully connected layer depth analysis}

To compensate for the reduced number of parameters in the Fourier, DWT, Nystr{\"o}mformer token-mixing mechanisms, we increase the depth of the fully connected (FC) layer after the feature extraction step in \ref{fig:dsnet}, using default ROI poolings and segment length $= [4, 8, 16, 32]$. The experimental results of varying FC layer depths on both SumMe and TVSum datasets are shown in Figure \ref{fig:accuracy_results}.



Softmax attention and Nystr{\"o}mformer attention show the most stable performance across FC depths on both datasets, suggesting robustness and reliability in varying configurations. Fourier and DWT token-mixing demonstrate greater sensitivity to FC depth changes, particularly on the SumMe dataset. This analysis indicates the importance of selecting appropriate attention mechanisms and FC depths to optimize model performance for specific datasets.

\section{Conclusion}
Traditional approaches for video summarization using transformer-based models often face computational challenges, especially with long video sequences. To overcome these limitations, we propose enhancement in DSNet by employing efficient token-mixing mechanisms such as Fourier, DWT, Nystr{\"o}mformer, optimized through anchor-based region proposals and varying pooling methods. Our experiments, conducted on the TVSum and SumMe datasets, show that our models achieve competitive F1 scores while significantly reducing GPU memory usage and parameter counts. The results highlight the stability and robustness of Nystr{\"o}mformer across varying Fully Connected (FC) layer depths, while Fourier and DWT token-mixing demonstrate sensitivity to these changes. We also see that while ROI pooling performs well on SumMe, FFT pooling consistently achieves the best results for TVSum, highlighting the importance of selecting the appropriate pooling method based on dataset characteristics for video summarization. Through comprehensive comparisons with existing state-of-the-art, we demonstrate that our approach offers a more computationally efficient alternative without compromising summarization accuracy.

\bibliographystyle{unsrt}
\bibliography{references}

\begin{thebibliography}{10}

\bibitem{statista_youtube_2024}
Statista.
\newblock Hours of video uploaded to youtube every minute as of 2023, 2024.
\newblock Accessed: 2024-09-12.

\bibitem{christel2006evaluation}
Michael~G Christel.
\newblock Evaluation and user studies with respect to video summarization and browsing.
\newblock {\em Multimedia Content Analysis, Management, and Retrieval 2006}, 6073:196--210, 2006.

\bibitem{rani2020social}
Seema Rani and Mukesh Kumar.
\newblock Social media video summarization using multi-visual features and kohnen's self organizing map.
\newblock {\em Information Processing \& Management}, 57(3):102190, 2020.

\bibitem{muhammad2020efficient}
Khan Muhammad, Tanveer Hussain, and Sung~Wook Baik.
\newblock Efficient cnn based summarization of surveillance videos for resource-constrained devices.
\newblock {\em Pattern Recognition Letters}, 130:370--375, 2020.

\bibitem{zhang2016context}
Shu Zhang, Yingying Zhu, and Amit~K Roy-Chowdhury.
\newblock Context-aware surveillance video summarization.
\newblock {\em IEEE Transactions on Image Processing}, 25(11):5469--5478, 2016.

\bibitem{vaswani2017attention}
Ashish Vaswani.
\newblock Attention is all you need.
\newblock {\em arXiv preprint arXiv:1706.03762}, 2017.

\bibitem{xiong2021nystromformer}
Yunyang Xiong, Zhanpeng Zeng, Rudrasis Chakraborty, Mingxing Tan, Glenn Fung, Yin Li, and Vikas Singh.
\newblock Nystr\"omformer: A nystr\"om-based algorithm for approximating self-attention, 2021.

\bibitem{lee2021fnet}
James Lee-Thorp, Joshua Ainslie, Ilya Eckstein, and Santiago Ontanon.
\newblock Fnet: Mixing tokens with fourier transforms.
\newblock {\em arXiv preprint arXiv:2105.03824}, 2021.

\bibitem{rochan2018video}
Mrigank Rochan, Linwei Ye, and Yang Wang.
\newblock Video summarization using fully convolutional sequence networks.
\newblock In {\em Proceedings of the European conference on computer vision (ECCV)}, pages 347--363, 2018.

\bibitem{fajtl2019summarizing}
Jiri Fajtl, Hajar~Sadeghi Sokeh, Vasileios Argyriou, Dorothy Monekosso, and Paolo Remagnino.
\newblock Summarizing videos with attention.
\newblock In {\em Computer Vision--ACCV 2018 Workshops: 14th Asian Conference on Computer Vision, Perth, Australia, December 2--6, 2018, Revised Selected Papers 14}, pages 39--54. Springer, 2019.

\bibitem{zhou2018deep}
Kaiyang Zhou, Yu~Qiao, and Tao Xiang.
\newblock Deep reinforcement learning for unsupervised video summarization with diversity-representativeness reward.
\newblock In {\em Proceedings of the AAAI conference on artificial intelligence}, volume~32, 2018.

\bibitem{feng2018extractive}
Litong Feng, Ziyin Li, Zhanghui Kuang, and Wei Zhang.
\newblock Extractive video summarizer with memory augmented neural networks.
\newblock In {\em Proceedings of the 26th ACM international conference on Multimedia}, pages 976--983, 2018.

\bibitem{wang2020linformer}
Sinong Wang, Belinda~Z Li, Madian Khabsa, Han Fang, and Hao Ma.
\newblock Linformer: Self-attention with linear complexity.
\newblock {\em arXiv preprint arXiv:2006.04768}, 2020.

\bibitem{choromanski2020rethinking}
Krzysztof Choromanski, Valerii Likhosherstov, David Dohan, Xingyou Song, Andreea Gane, Tamas Sarlos, Peter Hawkins, Jared Davis, Afroz Mohiuddin, Lukasz Kaiser, et~al.
\newblock Rethinking attention with performers.
\newblock {\em arXiv preprint arXiv:2009.14794}, 2020.

\bibitem{beltagy2020longformer}
Iz~Beltagy, Matthew~E Peters, and Arman Cohan.
\newblock Longformer: The long-document transformer.
\newblock {\em arXiv preprint arXiv:2004.05150}, 2020.

\bibitem{9706874}
Pranav Jeevan and Amit Sethi.
\newblock Resource-efficient hybrid x-formers for vision.
\newblock In {\em 2022 IEEE/CVF Winter Conference on Applications of Computer Vision (WACV)}, pages 3555--3563, 2022.

\bibitem{teutsch2015robust}
Michael Teutsch and Wolfgang Kruger.
\newblock Robust and fast detection of moving vehicles in aerial videos using sliding windows.
\newblock In {\em Proceedings of the IEEE conference on computer vision and pattern recognition workshops}, pages 26--34, 2015.

\bibitem{lian2022sliding}
Wenyi Lian and Wenjing Lian.
\newblock Sliding window recurrent network for efficient video super-resolution.
\newblock In {\em European Conference on Computer Vision}, pages 591--601. Springer, 2022.

\bibitem{yuan2009speeding}
Junsong Yuan, Zicheng Liu, Ying Wu, and Zhengyou Zhang.
\newblock Speeding up spatio-temporal sliding-window search for efficient event detection in crowded videos.
\newblock In {\em Proceedings of the 1st ACM international workshop on events in multimedia}, pages 3--8, 2009.

\bibitem{shou2017cdc}
Zheng Shou, Jonathan Chan, Alireza Zareian, Kazuyuki Miyazawa, and Shih-Fu Chang.
\newblock Cdc: Convolutional-de-convolutional networks for precise temporal action localization in untrimmed videos.
\newblock In {\em Proceedings of the IEEE conference on computer vision and pattern recognition}, pages 5734--5743, 2017.

\bibitem{wang2018segment}
Le~Wang, Xuhuan Duan, Qilin Zhang, Zhenxing Niu, Gang Hua, and Nanning Zheng.
\newblock Segment-tube: Spatio-temporal action localization in untrimmed videos with per-frame segmentation.
\newblock {\em Sensors}, 18(5):1657, 2018.

\bibitem{shou2016temporal}
Zheng Shou, Dongang Wang, and Shih-Fu Chang.
\newblock Temporal action localization in untrimmed videos via multi-stage cnns.
\newblock In {\em Proceedings of the IEEE conference on computer vision and pattern recognition}, pages 1049--1058, 2016.

\bibitem{jain2014action}
Mihir Jain, Jan Van~Gemert, Herv{\'e} J{\'e}gou, Patrick Bouthemy, and Cees~GM Snoek.
\newblock Action localization with tubelets from motion.
\newblock In {\em Proceedings of the IEEE conference on computer vision and pattern recognition}, pages 740--747, 2014.

\bibitem{yu2015fast}
Gang Yu and Junsong Yuan.
\newblock Fast action proposals for human action detection and search.
\newblock In {\em Proceedings of the IEEE conference on computer vision and pattern recognition}, pages 1302--1311, 2015.

\bibitem{escorcia2016daps}
Victor Escorcia, Fabian Caba~Heilbron, Juan~Carlos Niebles, and Bernard Ghanem.
\newblock Daps: Deep action proposals for action understanding.
\newblock In {\em Computer Vision--ECCV 2016: 14th European Conference, Amsterdam, The Netherlands, October 11-14, 2016, Proceedings, Part III 14}, pages 768--784. Springer, 2016.

\bibitem{zhu2020dsnet}
Wencheng Zhu, Jiwen Lu, Jiahao Li, and Jie Zhou.
\newblock Dsnet: A flexible detect-to-summarize network for video summarization.
\newblock {\em IEEE Transactions on Image Processing}, 30:948--962, 2020.

\bibitem{szegedy2015going}
Christian Szegedy, Wei Liu, Yangqing Jia, Pierre Sermanet, Scott Reed, Dragomir Anguelov, Dumitru Erhan, Vincent Vanhoucke, and Andrew Rabinovich.
\newblock Going deeper with convolutions.
\newblock In {\em Proceedings of the IEEE conference on computer vision and pattern recognition}, pages 1--9, 2015.

\bibitem{jeevan2022wavemix}
Pranav Jeevan, Kavitha Viswanathan, Amit Sethi, et~al.
\newblock Wavemix: A resource-efficient neural network for image analysis.
\newblock {\em arXiv preprint arXiv:2205.14375}, 2022.

\bibitem{ji2019video}
Zhong Ji, Kailin Xiong, Yanwei Pang, and Xuelong Li.
\newblock Video summarization with attention-based encoder--decoder networks.
\newblock {\em IEEE Transactions on Circuits and Systems for Video Technology}, 30(6):1709--1717, 2019.

\bibitem{apostolidis2021combining}
Evlampios Apostolidis, Georgios Balaouras, Vasileios Mezaris, and Ioannis Patras.
\newblock Combining global and local attention with positional encoding for video summarization.
\newblock In {\em 2021 IEEE international symposium on multimedia (ISM)}, pages 226--234. IEEE, 2021.

\bibitem{ghauri2021supervised}
Junaid~Ahmed Ghauri, Sherzod Hakimov, and Ralph Ewerth.
\newblock Supervised video summarization via multiple feature sets with parallel attention.
\newblock In {\em 2021 IEEE International Conference on Multimedia and Expo (ICME)}, pages 1--6s. IEEE, 2021.

\bibitem{zhang2018retrospective}
Ke~Zhang, Kristen Grauman, and Fei Sha.
\newblock Retrospective encoders for video summarization.
\newblock In {\em Proceedings of the European conference on computer vision (ECCV)}, pages 383--399, 2018.

\bibitem{potapov2014category}
Danila Potapov, Matthijs Douze, Zaid Harchaoui, and Cordelia Schmid.
\newblock Category-specific video summarization.
\newblock In {\em Computer Vision--ECCV 2014: 13th European Conference, Zurich, Switzerland, September 6-12, 2014, Proceedings, Part VI 13}, pages 540--555. Springer, 2014.

\bibitem{song2015tvsum}
Yale Song, Jordi Vallmitjana, Amanda Stent, and Alejandro Jaimes.
\newblock Tvsum: Summarizing web videos using titles.
\newblock In {\em Proceedings of the IEEE conference on computer vision and pattern recognition}, pages 5179--5187, 2015.

\bibitem{gygli2014creating}
Michael Gygli, Helmut Grabner, Hayko Riemenschneider, and Luc Van~Gool.
\newblock Creating summaries from user videos.
\newblock In {\em Computer Vision--ECCV 2014: 13th European Conference, Zurich, Switzerland, September 6-12, 2014, Proceedings, Part VII 13}, pages 505--520. Springer, 2014.

\bibitem{deng2009imagenet}
Jia Deng, Wei Dong, Richard Socher, Li-Jia Li, Kai Li, and Li~Fei-Fei.
\newblock Imagenet: A large-scale hierarchical image database.
\newblock In {\em 2009 IEEE conference on computer vision and pattern recognition}, pages 248--255. Ieee, 2009.

\end{thebibliography}

\end{document}